\documentclass[conference]{IEEEtran}
\IEEEoverridecommandlockouts
\usepackage{times}
\usepackage{epsfig}
\usepackage{cite}
\usepackage{amsmath,amssymb,amsfonts}
\usepackage{algorithmic}
\usepackage{graphicx}
\usepackage{textcomp}
\usepackage{xcolor}
\usepackage{amssymb}
\usepackage{rotating}
\begin{document}

\title{DeepGamble: Towards unlocking real-time player intelligence using multi-layer instance segmentation and attribute detection} %\\{\footnotesize Paper ID: 12}}

% \author{Danish Syed\thanks{\textit{Now at University of Michigan, dasyed@umich.edu}}
%     \textsuperscript{, \specificthanks{2}}\hspace{1cm}
% Naman Gandhi\thanks{Equal contribution}\hspace{1cm}
% Arushi Arora\thanks{Equal contribution}\hspace{1cm}
% Nilesh Kadam\hspace{1cm}
% Rasvan Dirlea\\
% ZS Associates\\
% {\tt\small \{danish.syed, naman.gandhi, arushi.arora, nilesh.kadam, rasvan.dirlea\}@zs.com}
% }

% \author{Danish Syed $^{* \dagger},$ Naman Gandhi $^{*},$ Arushi Arora $^{*},$ Nilesh Kadam \\
% {\tt\small dasyed@umich.edu \{naman.gandhi, arushi.arora, nilesh.kadam\}@zs.com}
% }

% \author{Danish Syed\thanks{equal contribution} \and Naman Gandhi\footnotemark[1] \and  Arushi Arora \and  Nilesh Kadam \\
% {\tt\small \{danish.syed, naman.gandhi, arushi.arora, nilesh.kadam\}@zs.com}
% }

% Naman Gandhi\thanks{Equal contribution}\hspace{1cm}
% Arushi Arora\thanks{Equal contribution}\hspace{1cm}
% Nilesh Kadam\hspace}{1cm}
% Rasvan Dirlea\\
% {\tt\small \{danish.syed, naman.gandhi, arushi.arora, nilesh.kadam, rasvan.dirlea\}@zs.com}
% }

% Ioannis Kansizoglou $^{1 *},$ Nicholas Santavas $^{1 *},$ Loukas Bampis $^{1},$ and Antonios Gasteratos $^{1}$
% Laboratory of Robotics and Automation, Democritus University of Thrace, Xanthi, Greece \{ikansizo, nsantava, lbampis, agaster $\} @$ pme.duth.gr

\author{\IEEEauthorblockN{Danish Syed\textsuperscript{* $\dagger$} \thanks{\newline * \textit{indicates equal contribution} \newline $\dagger$ \textit{this work was done when author was at ZS Associates}}}
\IEEEauthorblockA{
\textit{University of Michigan}\\
dasyed@umich.edu}
\and
\IEEEauthorblockN{Naman Gandhi\textsuperscript{*}}
\IEEEauthorblockA{
\textit{ZS Associates}\\
naman.gandhi@zs.com}
\and
\IEEEauthorblockN{Arushi Arora\textsuperscript{*}}
\IEEEauthorblockA{
\textit{ZS Associates}\\
arushi.arora@zs.com}
\and
\IEEEauthorblockN{Nilesh Kadam}
\IEEEauthorblockA{
\textit{ZS Associates}\\
nilesh.kadam@zs.com}

}

\maketitle

%%%%%%%%%%% Abstract %%%%%%%%%%%%%%
\begin{abstract}
Annually the gaming industry spends approximately \$15 billion in marketing reinvestment. However, this amount is spent without any consideration for the skill and luck of the player. For a casino, an unskilled player could fetch $\sim 4\times$ more revenue than a skilled player. This paper describes a video recognition system that is based on an extension of the Mask R-CNN model. Our system digitizes the game of blackjack by detecting cards and player bets in real time and processes decisions they took in order to create accurate player personas. Our proposed supervised learning approach consists of a specialized three-stage pipeline that takes images from two viewpoints of the casino table and does instance segmentation to generate masks on proposed regions of interests. These predicted masks along with derivative features are used to classify image attributes that are passed onto the next stage to assimilate the gameplay understanding. Our end-to-end model yields an accuracy of $\sim$95\% for main bet detection and $\sim$98\% for card detection in a controlled environment trained using transfer learning approach with 900 training examples. Our approach is generalizable and scalable and shows promising results in varied gaming scenarios and test data. Such granular level gathered data, helped in understanding player's deviation from optimal strategy and thereby separate the skill of the player from the luck of the game. Our system also assesses the likelihood of card counting by correlating player's betting pattern to the deck's scaled count. Such a system lets casinos flag fraudulent activity and calculate expected personalized profitability for each player and tailor their marketing reinvestment decisions.
\end{abstract}


%%%%%%%%%%% Introduction %%%%%%%%%%%%%%
\section{Introduction}

In this paper, we present \textit{DeepGamble}, a video recognition system that assesses players as they play a game of blackjack. To create personalized hold\% of any player, we need to understand 1) betting behavior/patterns, 2) player decisions vs. the optimal strategy, and 3) likelihood of fraud. A foundational step in this journey is to digitize the gameplay. Unlike slot machines where data per handle pull can be analyzed, table games cannot access this per hand play.

\textit{We propose a video recognition solution that digitizes the table gameplay. Proposed solution is built using a multi-stage machine learning pipeline that does instance segmentation (chips \& cards) along with predicting secondary image attributes (chip denomination, card face value, orientation and much more). This digitized information opens up multiple analytical avenues that can help creatively assess value and market a table game player.}

Proposed system was piloted with a live executive audience of an industry veteran entertainment client, where the executives were invited over in multiple sessions to play 15 hands of blackjack per player and the system recorded and analyzed the bets they made, the cards that were dealt to them, and the decisions they took. All this information was processed in real time. At the end of each session, the instant analysis results: average bet, game outcomes, blackjack skill level and evaluation of potential card counting were shared.

%-------------------------------------------------------------------------

%%%%%%%%%%% Motivation %%%%%%%%%%%%%%
\section{Motivation} \label{motivation}
When a player games at a slot machine, the casino has perfect information about that player's wager, the number of handle-pulls, length of play. However, if a player prefers to play a table game, the casino must use guesswork from dealers and supervisors to estimate the player’s value to the casino - which may be off by huge margins (as high as 90\%). This margin of error can significantly distort the level of marketing reinvestment which the player receives in the form of free play, giveaways, comps. Overestimating a player’s value can destroy the lifetime profitability of the casino and underestimating it can negatively impact the player’s loyalty. 

While accurate main bet wager valuation remains a priority, not all bets are equal, and side bet hold\% can be $10\times$ main bet. To exemplify, for a basic blackjack having main bet strategy only, casinos have a hold\% of 0.5\% while on a buster blackjack with side bet \cite{proc-wesley}, casino's hold can go as high as 6\%.

\begin{figure}[h!]
\begin{center}
   \includegraphics[width=1\linewidth]{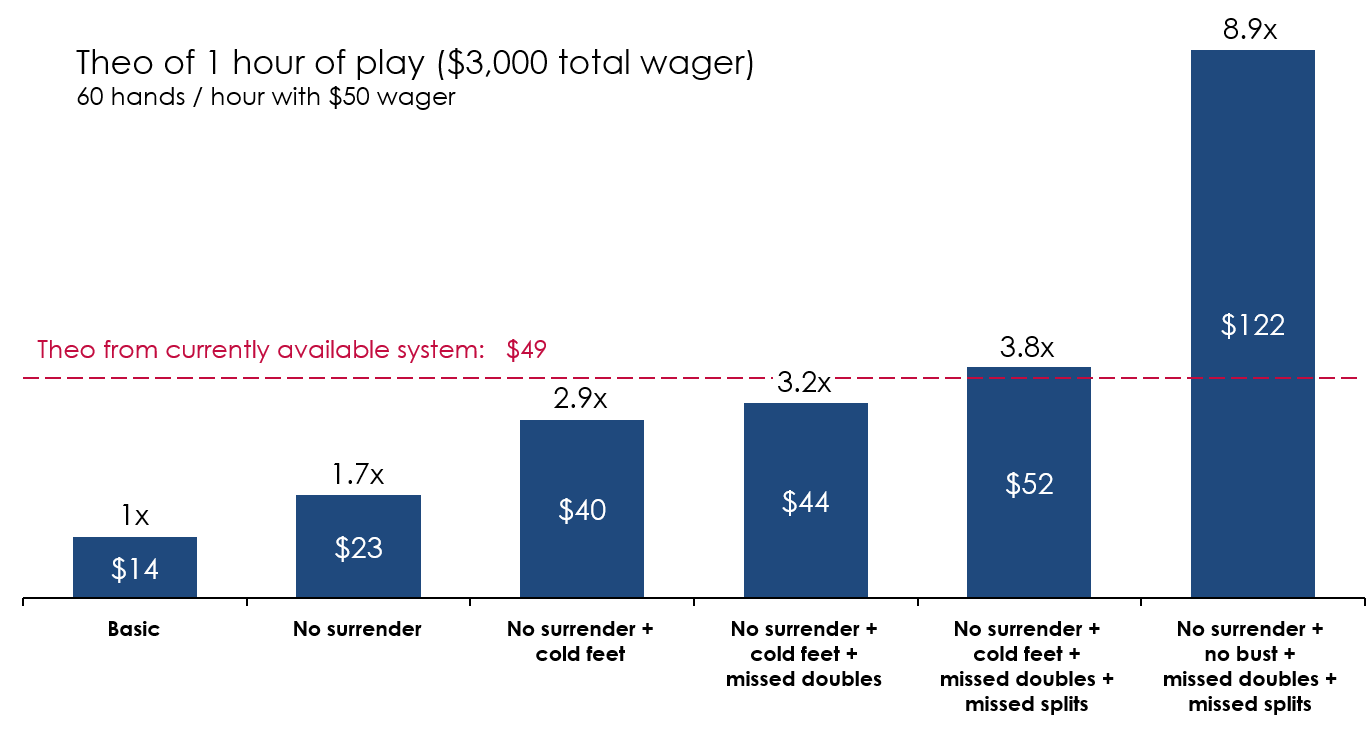}
\end{center}
   \caption{Impact of various blackjack strategies on a player's theo compared with currently available systems.}
\label{fig:theo-strategy}
%\label{fig:long}
%\label{fig:onecol}
\end{figure}

In addition to wager size and bet type, player decisions also profoundly impact player valuation. To quantify it, we ran Monte-Carlo simulations by extending the work of Seblau's blackjack simulator~\cite{git-seblau} to understand the impact of player decisions on theo\footnote{Theo is the theoretical value that a player will lose if they played for a certain amount of time with a certain number of bets in a game.} they generate. With game rules and strategies as specified in \cite{web-bbjstudy}, we simulated 60 hands/hour with \$50 wagers in each hand and understood the impact on theo by varying the gameplay strategies. If currently available hardware systems were to evaluate such a player, such a system would state this player as \$49 average bet player, while in reality player's decisions altered the casino hold and taking suboptimal strategies increases the theo by at max by $\sim8.9\times$ which makes this suboptimal player more profitable to the casinos. The impact of different blackjack strategies on the player's theo can be seen in Figure \ref{fig:theo-strategy}. Thus, effective gameplay valuation will help casinos calculate a personalized hold\% based on the player’s observed strategy (skill) and segment players by skill level and ultimately separate skill from luck.

\section{Related Work}
Computer vision gives an ideal, simple, fast, and inexpensive solution. However, there is little-published work in the direction of unlocking player intelligence. Notably, since the early 2000s, we see some advancements as we see a combination of hardware and software used to solve this. Below sections describe the efforts taken by the industry. 

\subsection{RFID Based Methods}
Recent efforts from large entertainment houses are focused on radio-frequency identification tags (RFID) embedded in chips \cite{pat-rfid-1, pat-rfid-2}. We also did an empirical study at our client's location to assess player's value using their existing RFID setup as described in the Section \ref{motivation}. Our findings suggested that even hardware based systems have a 2\% error (Figure \ref{fig:theo-strategy}). Despite high accuracy, these solutions are not as prevalent because they are prohibitively expensive and susceptible to fraud and table-failure. Due to the blind broadcast nature of RFID, these systems have been plagued with security concerns that players would be able to broadcast a compromised signal, possibly representing different chip denominations. While encryption within the chips can provide an extra layer of security, this dramatically increases the cost of each chip.

\subsection{Computer Vision Based Methods}
Clear Deal \cite{proc-wesley} used a combination of line detection, corner detection and template matching to detect the value of the cards as they are dealt throughout the game. The system analyzed the quality of the shuffle carried out by the dealer by comparing the deal across hands and detected card counting by monitoring game decisions and comparing them with basic strategy. However, this system had no way of monitoring the size or variation of bets placed by the player.

Zaworka \cite{proc-zaworka} tracked a Blackjack game by detecting cards and players’ chip stacks as they bet, in real time. Overall accuracy was 97.5\% for detecting playing cards and chip stacks, even with occlusion. However, the system only detected the presence, not the values of cards and chip stacks. The system used an electronic chip tray, which is an additional investment across the enterprise. Template matching and a combination of heuristics were used by Zheng \cite{proc-zheng} to match cards invariant to rotation, but the technique did not handle face cards well, did not model chips or bet sizes, and did not produce a final usable system. 

Generic object recognition has seen much use in the past decade \cite{book-ponce}. There are a number of discriminative approaches proposed, perhaps the most common of which is the use of invariant features \cite{art-kryst}, most notably the scale invariant feature transform (SIFT) \cite{art-sift}. 

\begin{figure*}
\begin{center}
% \fbox{\rule{0pt}{2in} \rule{.9\linewidth}{0pt}}
\includegraphics[width=0.9\linewidth]{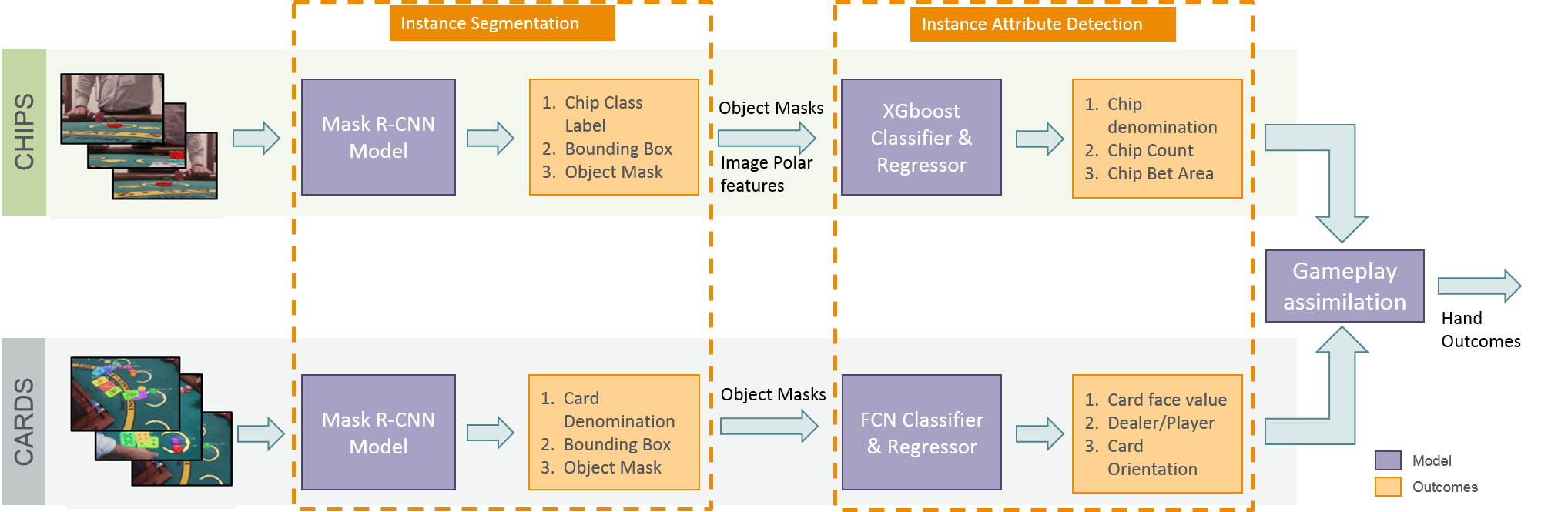}
\end{center}
   \caption{DeepGamble Model Pipeline. The instance segmentation pipelines have two streams - chips and cards detection. Instance attribute detection module for chips uses XGBoost, and for cards, a single FCNN is used. Results are used for gameplay assimilation.}
\label{fig:model-pipeline}
\end{figure*}

\subsection{Other Commercially Available Methods}
There are a few commercial attempts to market systems for card counting monitoring. Tangam Gaming \cite{web-tangam} produces an automated card recognition system that requires the use of specialized hardware such as RFID. The MindPlay21 system relied on a range of specialized hardware which included 14 cameras, invisible ink, and RFID tags. Cameras were used to scan the cards as they were dealt, as each card had been marked with a unique barcode painted in special ink \cite{web-terdiman}. The cost of \$20,000 per table, the unreliable components and the slow speed of operation led to the company going out of business in 2005. There are many patents also suitably guarding this space thereby inhibiting a large-scale commercial solution \cite{pat-ctm, pat-cshfl, pat-gcrs}. In the 2017 edition of the G2E conference, VizExplorer \cite{web-viz} launched its new tableViz\textsuperscript{TM} with ChipVue\textsuperscript{TM} product, and this solution provided reliable bet recognition data for a few table games. tableViz\textsuperscript{TM} system struggled to maintain high-accuracy and wasn't developed actively, whereas ChipVue\textsuperscript{TM} product released a new version in 2019 edition of G2E conference \cite{web-viz19} that continues to just provide accurate bet assessment.

\subsection{Comparison to our System}
The maturity of solutions in terms of accurate card detection, coupled with building a gameplay understanding is still in its nascent stages for the past two decades.

By tailgating on the recent advancements in Deep learning focused with attention to the Convolution neural networks to solve the problem of assimilating strategic gameplay information at the player level, we propose a system called \textit{DeepGamble} with the reference solution architecture (Refer Figure \ref{fig:model-pipeline}). 

%%%%%%%%%%% \textit{DeepGamble} %%%%%%%%%%%%%%
\section{DeepGamble}

In this section, we outline all the components of the \textit{DeepGamble} system. Our system understands the hand outcomes of a game of blackjack by understanding multitude of objects placed on the table and their spatial relationships in three sequential steps 1) Instance Segmentation, 2) Instance Attribute Detection, and 3) Gameplay Assimilation. 

\subsection{Instance Segmentation}
To build the gameplay understanding, we focus on detecting via two vantage points - chipboard viewpoint \textit{(using Chip detection pipeline)} and overhead viewpoint \textit{(using Card detection pipeline)} as stated in the Figure \ref{fig:model-pipeline}. This task of segmentation is accomplished using Mask R-CNN - an instance segmentation network. We refer the readers to the to the original publication for architectural and technical details \cite{art-maskrcnn}.

Our system segregated the detection of chips and cards via two separate models as through a single viewpoint we cannot capture the more delicate nuances of the objects. To elaborate, from the chipboard viewpoint, we cannot necessarily see all the cards that are kept close to the players, whereas, from the overhead viewpoint, we do not see the depth of the main bet chips for the players. Hence trade-off of multiple models vs. accuracy of detection led us down the path of separate models as our design principle.

We did domain-based fine-tuning for the two separate Mask R-CNN models prepared for chip and card dataset to identify the chip stacks with their denomination and cards with their face-value respectively, as shown in Figure \ref{fig:model-pipeline}. \\

\textbf{Training instance segmentation models:} We trained two separate Mask R-CNN models: 1) chip detection model, and 2) card detection model by using the weights of the model pre-trained on COCO dataset \cite{proc-coco, web-matterport}. The Mask R-CNN models were trained using Keras 2.1 functional API \cite{web-keras} running on Tensorflow backend \cite{web-tf} with NVIDIA CuDNN GPU acceleration libraries \cite{art-cudnn}. We used SGD algorithm to train 80 epochs (stopped using early stopping and scheduled with learning rate scheduler that uses an inverse time decay schedule with initial value of 0.01) for all images in the training set. We used four P100 Nvidia GPUs to train the model with the throughput of two images per GPU in a batch, thus effectively using a batch size of 8, and also included the combination of early stopping and model checkpoint callbacks to reduce overfitting. 

Additionally, to make the training more robust, we performed real-time data augmentation to the image using Imgaug \cite{web-imgaug}, where each image was randomly applied one of the transformations: contrast normalization, gaussian blur, rotation, translation, or shear. 

\subsection{Instance Attribute Detection}
Post instance identification, predicted masks and derivative features are passed on to their respective second layer models for further instance attribute prediction. We predicted the chip count of stack, the card orientation and the relative placement of objects on the table.

\textbf{Secondary attributes for chip detection module:} We used eXtreme Gradient Boosting (XGBoost) algorithm \cite{proc-xgboost} for counting the number of chips in a stack and classifying the bet placement area \textit{(main bet, side bet, or others)}. Both the models expected a set of derived features from the first stage model - namely width and height of the chip stack and relative polar coordinates (radius, theta) as highlighted in Figure \ref{fig:model-pipeline}, these features were calculated using the masks generated from the first stage model.\\

\textbf{Secondary attributes for card detection module:} We created a six layer Fully Convolution Neural Network (FCNN) from ground up for detecting the hand (player(s) or dealer), the orientation of cards (vertical/horizontal) and placement area of chips (main bet/side bet). Detection of placement of chips in the bet areas is far more vivid and clear from the top vantage point as opposed to from the chipboard level.

\begin{figure*}
\begin{center}
% \fbox{\rule{0pt}{2in} \rule{.9\linewidth}{0pt}}
\includegraphics[width=0.8\linewidth]{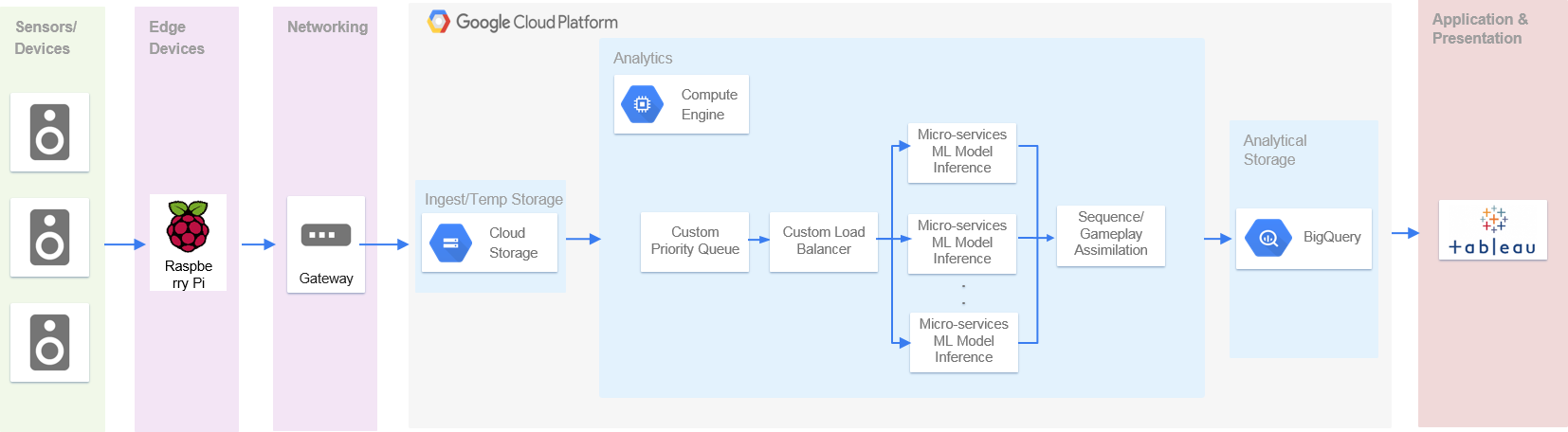}
\end{center}
   \caption{DeepGamble System Architecture. High-resolution cameras, Raspberry Pis are connected via a gateway to the Google Cloud Platform where inference models are deployed as micro-services to perform inference in real-time. After assimilating the game play, results are pushed to Google BigQuery for further analysis and real-time dashboards are generated.}
\label{fig:architecture}
\end{figure*}

\subsection{Gameplay Assimilation}
Both Instance segmentation and Instance attribute detection provided a frame level analysis of the objects - bets placed and cards dealt. It became imperative to create a unified view of the hand from individual frame level results to understand the gameplay and evaluate the player decisions. This step integrated the outputs of the chip and card models from previous steps and stitched the predicted images to create a retrospective hand replay video. 

\textbf{Chips:} To estimate the player's initial bet value, we majority voted on the predicted value of the main bet after we have received at least 70\% of the frames for each player (to factor in for the fact that the images might be lost in transit).

\textbf{Cards:} Unlike chips, where we work with most frames, for cards we need all frames of the hand to start the sequence assimilation as the order of the cards dealt serves as an essential feature in assessing player's skill. Per the rules of the game, we know that each player will be initially dealt two cards in a circular fashion until the dealer's first card is revealed.  

A tricky situation arises as previous cards dealt to the players become partially obstructed upon placement of a new card, so we only have a few frames window where we observe the cards unobstructed. Hence, we majority voted on the cards detected in each frame in its window of unobstructed view. The trigger to majority vote happened only at state change of cards on the table.

Post the first two cards, the gameplay progression can take multiple paths as players have an intervention and a decision to make. At this point, we evaluate each player's decision and strategy.

The gameplay naturally shifts to the dealer after the dealer's hole card is revealed and they start to draw their cards. We used the spatial features of the bounding box to detect the sequence of the dealer cards as they are generally translated on the axis to accommodate a new card being dealt. The hole card is opened to the left of the first card. Dealer's third card and others are dealt to the right of the first card. Due to the overhead camera position, this spatial distance decreased as we moved from left to right thus enabling us to identify the correct card. Per the table game, usually the dealer stands on 17s, and no further cards are dealt. At this point, we evaluated the game outcomes and assessed the net win/loss amount. (Evaluation detailed in Section \ref{gameplay-eval})

%%%%%%%%%%% System Architecture %%%%%%%%%%%%%%
\section{System Architecture}

We now describe the system architecture of our solution \textit{DeepGamble}. Our solution is created and deployed on scalable Google Cloud Platform and is designed to operate in real time. Figure \ref{fig:architecture} outlines a high-level system architecture for \textit{DeepGamble}. 
The blackjack table was outfitted with small, inexpensive computing devices (raspberry pi), high-resolution cameras, and other supporting hardware. The cameras were secured on the table by housing them in 3D printed holders. 

The workflow of the \textit{DeepGamble} system consists of the following steps as shown in Figure \ref{fig:architecture}:
\begin{enumerate}
    \item The cameras connected to the Raspberry Pis capture the images at 10 fps.
    
    \item The images from all sensors are brought to local network storage, and from there these images are uploaded to Google Cloud Storage. 
    
    \item The pointer reference to these images are pushed in a priority queue to make sure that even if an image from the previous hand arrived late due to network issues, it gets processed first. 
    
    \item The inference models are deployed as micro-services which pulls the enqueued pictures with the help of a custom load balancer and process them parallelly in an asynchronous manner.
    
    \item Each frame as it gets predicted is pushed to Google BigQuery and post completion of a hand the frames are assimilated to create a unified understanding of the hand.
    
    \item The application layer accesses the consolidated hands' data from BigQuery to power our proposed applications.
\end{enumerate}

The asynchronous, parallel and scalable nature of our cloud-based architecture enable us to generate real-time inferences.

%%%%%%%%%%% EXPERIMENTS %%%%%%%%%%%%%%
\section{Experiments}
In this section, we present an extensive evaluation of \textit{DeepGamble}. First, we describe the dataset used for training and evaluation. Then, we discuss  metrics used for evaluating each component and overall pipeline of \textit{DeepGamble}. Finally, we share the performance of \textit{DeepGamble} system on holdout dataset.

\begin{figure}[h!]
\begin{center}
   \includegraphics[width=0.9\linewidth]{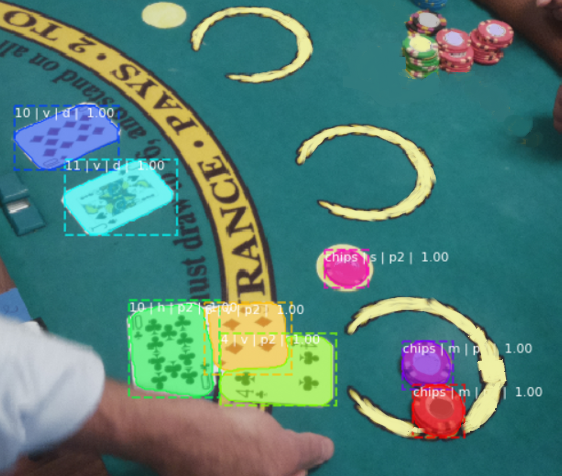}
\end{center}
   \caption{Annotation example for overhead POV. Ten of spades is annotated as - \{10, h, p2\} - card denomination, card orientation, card location (player 2). Side bet is annotated as - \{chips, s, p2\} - chip flag, chip stack bet area, chip stack location}
\label{fig:annotation}
\end{figure}

\subsection{Dataset} \label{val-data}
The open-source labeled datasets did not contain our domain-specific data elements i.e, chips and cards. Hence we resorted to using transfer learning. We leveraged the COCO public dataset \cite{proc-coco} for object detection and segmentation. We captured and annotated $\sim$900 images from our two different points of views - chipboard and overhead POV. We used open-source labelme \cite{art-labelme} software to annotate polygons for the objects in each frame.\\

\textbf{Chipboard POV:} Two cameras were installed on the table to capture the bet area of their corresponding player with a resolution of $2000*1006$ pixels. Each object polygon was assigned a comma separated string label with three attributes: 1) Chip stack color (red, green, blue, or black), 2) Chip stack bet area (main, side, or others) and 3) Chip stack count.\\

\textbf{Overhead POV:} The camera installed on the tower captures the whole table from the top with the same resolution as chipboard view. The polygon annotation string also had three attributes: 1) Card denomination (1 to 13) OR a chip flag in case of chip stack, 2) Card orientation (horizontal or vertical) or chip stack bet area (main or side) and 3) Card or chip stack location (player or dealer). 
These attributes were extracted from the string while training, an example of the annotation is shown in the Figure \ref{fig:annotation}.\\

\textbf{Pipeline validation dataset:} To evaluate the performance of our two independent pipelines (Chip detection \& Card detection as shown in the Figure \ref{fig:model-pipeline}) we collected images based on a predefined set of object combinations.

For Chip detection pipeline, we captured 1200 images with all the possible combinations of different denomination chips (red, green, blue \& black). By making an exhaustive validation dataset, we were able to find various nuances in the model and training. \textit{One example} could be the misclassification of black chips as red due to the high frequency of red chips in the training data. Also, the dim lighting of the casino floor confuses the red chips with black.

Similarly, for Cards pipeline, we recorded 150 hands ($\sim$70 images per hand) translating to 600 unique gameplay scenarios with each card value in obstructed and unobstructed positions. The table sizes were varied which caused a few player cards to be misclassified as dealer cards and vice-versa. As the misclassification generally happened post the fourth card was dealt to a player/dealer, we used the nature of the game and dealing to allocate the cards to the player or dealer correctly.\\

\subsection{Evaluation \& Results}

\begin{figure*}
\begin{center}
% \fbox{\rule{0pt}{2in} \rule{.9\linewidth}{0pt}}
\includegraphics[width=0.8\linewidth]{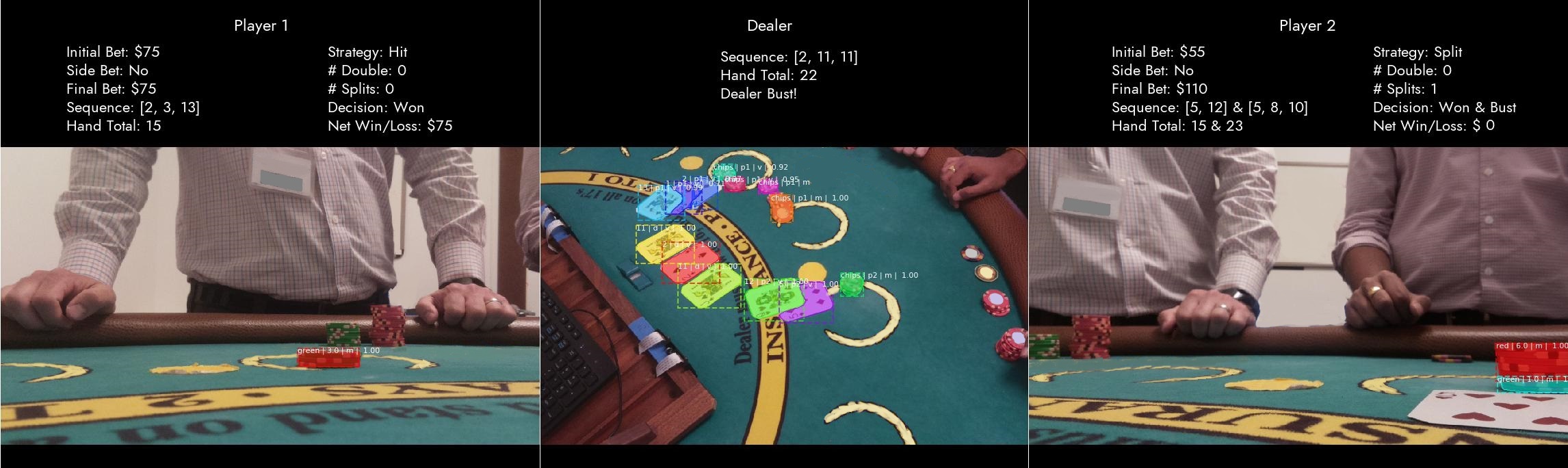}
\end{center}
   \caption{Gameplay Evaluation: Player's bets, card sequence, initial strategy and outcome of the player(s) \& dealer are evaluated.}
\label{fig:game-eval}
\end{figure*}

We will now look at the evaluation of the individual components of the model pipeline and then move onto the overall evaluation of the two independent sub-pipelines(chip and card).
Since there is very less published work on table gaming systems which are capable of digitizing the whole game of blackjack, we relied on our industry understanding to evaluate our model. However, we compared our chip and card detection pipeline with the work done by Krists \textit{et al}. \cite{art-bjpaper} which was evaluated in a constrained setup.\\

\textbf{Evaluation metric of instance segmentation:} For instance segmentation task, the standard metric for evaluation is a multi-task loss on each sampled ROI. It is defined as: 

\begin{equation} \label{eq:1}
    L = L_{cls} + L_{box} + L_{mask}
\end{equation}

The classification loss \textit{L\textsubscript{cls}}, bounding-box loss \textit{L\textsubscript{box}} and mask loss \textit{L\textsubscript{mask}} are identical as those defined in \cite{art-maskrcnn}.\\

\textbf{Evaluation metrics for instance attribute detection:} We predicted four attributes from the second layer models, two from both the pipelines consisting of classification and regression. For chip stack bet area, card location and card orientation we used multi-class logarithmic loss, and for the chip stack count, we used root mean squared error (RMSE) to evaluate their performance.

After tuning each component at the micro level, we tested the performance of the two pipelines on our synthetic validation dataset. We devised metrics of business significance that evaluated the performance of the pipelines as a whole.\\

\textbf{Evaluation of Chip detection pipeline:} To holistically evaluate the Chip detection pipeline, we used the chip evaluation data set as mentioned in Section \ref{val-data}. We evaluated the main bet accuracy for the overall bet and also across individual color in the stack as shown in Table \ref{bet-acc}. Krists \textit{et al}. \cite{art-bjpaper} used a BumbleBee2 stereo camera to calculate the size of the bet. Even though they were able to achieve 99\% accuracy on a small dataset, their chip setup was limited to a single color stack whereas our model achieved 95\% bet color detection accuracy on a realistic setup (with multiple color stacks in main bet area).

\begin{table}[h!]
\caption{Bet Detection Accuracy}
\begin{center}
\begin{tabular}{|c|c|}
\hline
Main Bet \textit{(overall \& color wise)} & Accuracy\\
\hline\hline
\textbf{Overall Stack} & \textbf{95\%}\\
Red Stack & 98.83\%\\
Green Stack & 99.6\%\\
Blue Stack & 100\%\\
Black Stack & 99.5\%\\
\hline
\end{tabular}
\end{center}
\label{bet-acc}
\end{table}

Although our bet accuracy results were satisfactory, we wanted to understand the effectiveness of our chip detection pipeline in terms of bet size calculation. Hence, we further evaluated the mean absolute percentage error (MAPE) of main bet chip count and value-weighted chip count (since an error in a higher denomination chips would be less desirable) as shown in Table \ref{bet-mape}. Our final model had an error of 7.35\% on value weighted chip count (same as total bet value), we also believe that our chip evaluation can act as a strong baseline to benchmark progress.

\begin{table}[h!]
\caption{Bet Value MAPE}
\begin{center}
\begin{tabular}{|c|c|}
\hline
Main Bet\textit{(Count \& Value weighted)} & MAPE\\
\hline\hline
\textbf{Chip count} & \textbf{7.12\%}\\
Value weighted chip count & 7.35\%\\
\hline
\end{tabular}
\end{center}
\label{bet-mape}
\end{table}

\textbf{Evaluation of Card detection pipeline:} For cards, we were interested in identifying the card outlines, the accuracy of the card position, orientation and the face value of the topmost unobstructed card only as we always had few frames in which a card was visible with no obstructions. To evaluate the card pipeline, we used the card evaluation dataset as mentioned in \ref{val-data}. We looked at the accuracy of identifying the card outline, the unobstructed card value - separate by face cards and non face cards (value$\leq$10). For face cards - J, Q, K - as long as the face card was identified, it was marked as accurate as these cards hold the same numerical value of 10 in blackjack and do not affect the game outcome. It was essential to correctly identify the player/dealer cards and the orientation (horizontal or vertical) of the card as these affect the outcome of the hand - Win/Loss/Stand/Push. The accuracy numbers for 150 hands are listed in Table \ref{card-acc}.

\begin{table}[h!]
\caption{Card detection accuracy for different components of DeepGamble and the model proposed by Krists \textit{et al}.\cite{art-bjpaper}.}
\begin{center}
\begin{tabular}{|c|c|c|}
\hline
Accuracy Metrics &  DeepGamble & Krists \textit{et al}.\\
\hline\hline
Single Card Outline & 100\% & 100\%\\
Multiple Card Outline & 100\% & 100\%\\
\hline
Overall Card Value & 98\% & \textbf{99.75\%}\\
Face Card Value & 100\% & --\\
Non-face Card Value & 96\% & --\\
\hline
Player Card Position & 99\% & --\\
Dealer Card Position & 98\% & --\\
Card Orientation(h or v) & 95\% & --\\
\hline
\end{tabular}
\end{center}

\label{card-acc}
\end{table}

In a parallel system \cite{art-bjpaper}, tested under controlled environment - with only 400 images; essentially testing each card 10 times - the author reported a card face value accuracy of 99.75\%. Whereas our proposed system, which was tested in near-real life scenario - with multiple combinations of actual players and dealers achieved a  comparable accuracy of card face value of 98\% and in few nuances even higher. Currently, there are no empirical benchmarks for evaluating the additional accuracy metrics like - side bets, card positions, orientation and game outcome.

For the side bets, we considered a \textit{boolean} flag for whether a side bet was placed or not. We were 98\% accurate in detecting the side bet for players.

%%%%%%%%%%% Applications %%%%%%%%%%%%%%
\section{Applications}
In this section, we will discuss the application stack built upon the computer vision pipeline. Modules in application stack work in parallel and  leverage the predictions of the hands stored in Google BigQuery to create analytical dashboards.

\begin{figure*}
\begin{center}
% \fbox{\rule{0pt}{2in} \rule{.9\linewidth}{0pt}}
\includegraphics[width=0.9\linewidth]{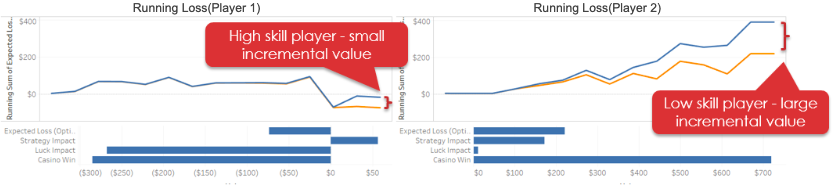}
\end{center}
   \caption{Player skill dashboard: The orange line is the optimal strategy, and the blue line is the player's actual strategy. As Player 1 follows the optimal strategy for most hands and deviates very less from it, he is not penalized for his strategy, but as he had better luck, he ends up with an overall win. Player 2 deviates from the optimal strategy, is penalized for his low skill and thus ends up offering the casino a large incremental value from the expected loss.}
\label{fig:dash-skill}
\end{figure*}

\subsection{Gameplay Evaluation} \label{gameplay-eval}
A player has to take a strategic decision post two cards are dealt to them. The decision could be whether to hit for another card, stand on his current hand, double his bets for the third card or split the hand and play two hands. We will now describe the markers we used to identify the player's strategy:

\begin{itemize}
    \item \textbf{Hit} - If the third card dealt to the player is placed in the vertical orientation.

    \item \textbf{Double Down} - If the third card dealt to the player is placed in the horizontal orientation, and there were two chip stacks in the main bet area.
    
    \item \textbf{Split} - A player is only allowed to split if their first two cards are of the same face value. They also placed another chip stack in the main bet area, and both cards were visible in the unobstructed view.   
    
    \item \textbf{Stand} - In case the player decided to stand on their current hand, they will not be dealt a third card.
\end{itemize}

The aim is to reach closer to the total of 21 on your hands, without going over and making sure the player's total is higher than the dealer's hand total. Usually, the dealer must draw to a total of 17. Screenshot of the evaluated gameplay - hand outcome, strategy and net win/loss of a hand is shown in Figure \ref{fig:game-eval}. The hand outcome based on the dealer and player hand's total is shown in Table \ref{game-outcome}.

\begin{table}[h!]
\begin{center}
\caption{Outcome based on player and dealer hand total}
\begin{tabular}{|c|c|}
\hline
Total & Player Outcome\\
\hline\hline
Player  $>$ 21 & Bust \\
Player $>$ Dealer & Won \\
Player = Dealer & Push\\
Player $<$ Dealer & Loss\\
Dealer  $>$ 21 and Player $\leq$ 21 & Won \\
Player Hits 21 on First 2 cards & Blackjack \\
\hline
\end{tabular}
\end{center}
\label{game-outcome}
\end{table}

\subsection{Skill Evaluation}
An ideal or basic strategy is defined using the player's first two cards and dealer's face card. Such a strategy dictates that given a situation, should we hit, stand or double-down. This ideal strategy can be derived by simulating a large number of hands. These set of rules or strategy were simulated using the Seblau’s blackjack simulator \cite{git-seblau} to find the optimal pathway. The skill level of a player could be assessed based on their deviation from this ideal strategy. Adherence to the ideal strategy will deem the player adroit. Figure \ref{fig:dash-skill} compares two players and their skill.

Skill differences dramatically impact player's value that’s why high-rollers are closely monitored by the floor supervisors for more personalized evaluation. However, today the same hold is assumed for all the ordinary players since there is no actionable and scalable intelligence.

\subsection{Card Counting Detection}
The basic theory behind card counting is simple – the player is at an advantage when more face cards are remaining in the deck because the dealer is more likely to bust. A player can take advantage of this fact by keeping track of the cards that have come out of the shoe and adjusting their bet accordingly.

The most common system of card-counting is called the \textit{“high-low”} system \cite{web-highlow}. Players determine their bet size using the \textit{scaled count} \cite{web-highlow} at the start of each hand, betting more on hands that start with a higher scaled count to lower the house advantage. To determine whether a player is counting cards, we have to look at the relationship between the scaled count$(x)$ and the hold percent$(y)$ of a casino at any given point. This was achieved by using the Seblau’s blackjack simulator \cite{git-seblau}, we generated the data for the variation in scaled count$(x)$ and its effect on the hold percentage$(y)$. It was an extremely close linear fit$(R^2 = 0.98326)$. Using the linear relationship in Eq. \ref{eq:5}, we can calculate the expected hold of every hand given the scaled count$(x)$. 
\begin{equation} \label{eq:5}
    y = -0.00474x + 0.00512 
\end{equation}

\begin{figure}[h!]
\begin{center}
   %\includegraphics[width=0.8\linewidth]{egfigure.eps}
   \includegraphics[width=0.8\linewidth]{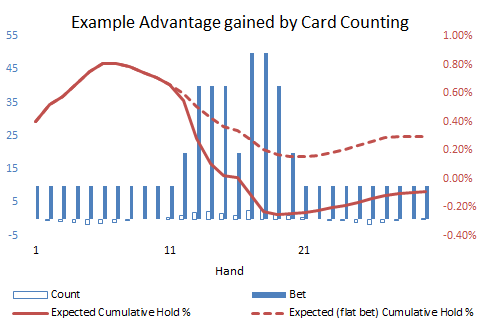}
\end{center}
   \caption{The gap between these two lines isolates the advantage the player created by tracking the count and betting accordingly. In this particular case, the card counting and bet variation shifts the advantage for the house to a slight advantage for the player.}
\label{fig:dash-cardcount}
\end{figure}

\textbf{Testing the card counting module:} We designed an experiment where a player counted the cards and placed the bet higher as the scaled count rose in the player's favor. We compared this player's trajectory with a flat bet player's journey (a constant bet for every hand) to evaluate the competitive advantage gained as shown in Figure \ref{fig:dash-cardcount}.

The cumulative expected hold percent($H_{player}$) and flat bet cumulative hold percent($H_{flatbet}$) for $N$ hands are calculated using Eq. \ref{eq:6} \& Eq. \ref{eq:7} respectively.

\begin{equation} \label{eq:6}
    H_{player} =
    \frac{\sum_{i=0}^{N} Bet_i * Expected Hold_i}{Total  Amount Bet}
\end{equation} 

\begin{equation} \label{eq:7}
    H_{flatbet} =
    \frac{AverageBet*\sum_{i=0}^{N} Expected Hold_i}{Total  Amount Bet}
\end{equation} 

%%%%%%%%%%% Conclusion %%%%%%%%%%%%%%
\section{Conclusion}
In this work, we presented a Mask R-CNN based approach for a new domain of assessing the player's worth as they play a game of blackjack. Our method, \textit{DeepGamble}, takes images from two viewpoints - chipboard and overhead, to predict average bet, game outcome, blackjack skill level and the likelihood of card counting in real time. The proposed method can easily adapt to different blackjack tables with different payout rules. Minimal fine-tuning might be needed as we change scales and perspectives. Extensive experiments on multiple hands ($\sim 150$) demonstrated the efficiency and effectiveness of the proposed method over the state-of-the-art. 

We do understand that as dealing style varies from dealer to dealer, we may have a lot of occlusion in the frames due to dealer's hand, which currently our model is able to discard as fewer feature maps in Mask R-CNN fire on hands of different sizes. In future work, we want to directly embed the occlusion of hand as a weighting layer into Mask R-CNN. It will produce different weights to combine feature map outputs at every pixel depending on the occlusion.

\section*{Acknowledgment}
The authors would like to thank Arun Shastri, Rasvan Dirlea, Mike Francis, Akshat Rajvanshi, Manoj Bheemineni, Brendan Riley, Geoff Cohn, Jayendu Sharma, Thompson Nguyen and others who contributed, supported, guided and collaborated with us during the development and deployment of our system.

% \printbibliography
% \bibliography{main}
% \bibliographystyle{ieee_fullname}
% \input{main.bbl}

{\small
\bibliographystyle{IEEEtran}
\bibliography{main}
}

% {\small
% \bibliographystyle{ieee_fullname}
% \bibliography{egbib.bib}
% }

\end{document}